\documentclass[final]{cvpr}

\usepackage{times}
\usepackage{epsfig}
\usepackage{graphicx}
\usepackage{amsmath}
\usepackage{amssymb}

\graphicspath{{fig/}}
\usepackage{float}

\usepackage[pagebackref=true,breaklinks=true,colorlinks,bookmarks=false]{hyperref}

\def\cvprPaperID{2563} 
\def\confYear{CVPR 2021}

\begin{document}

%
%

\title{BCNN:  A Binary CNN With All Matrix Ops Quantized To 1 Bit Precision}

\author{Arthur J. Redfern \qquad Lijun Zhu\\
Texas Instruments, Inc\\
12500 TI Boulevard, Dallas TX 75243\\
{\tt\small \{redfern, lijun\}@ti.com}
\and
Molly K. Newquist\\
Georgia Institute of Technology\\
North Avenue NW, Atlanta, GA 30332\\
{\tt\small mnewquist3@gatech.edu}
}

\maketitle

%
%

\begin{abstract}
This paper describes a CNN where all CNN style 2D convolution operations that lower to matrix matrix multiplication are fully binary.  The network is derived from a common building block structure that is consistent with a constructive proof outline showing that binary neural networks are universal function approximators.  71.24\% top 1 accuracy on the 2012 ImageNet validation set was achieved with a 2 step training procedure and implementation strategies optimized for binary operands are provided.
\end{abstract}

%
%

\section{Introduction}

CNNs have achieved wide spread success in vision applications.  The feature encoder portion of an image classification network typically forms the backbone of networks designed for semantic segmentation, object detection, object segmentation, depth estimation and motion estimation.  Minimizing the complexity of the backbone while achieving a high level of accuracy is a key to enabling the use of CNNs in resource constrained environments.

The complexity of CNNs is dominated by the CNN style 2D convolution operator.  This operator takes a 3D input tensor and a 4D weight tensor and produces a 3D output tensor.  CNN style 2D convolution can be lowered to matrix matrix multiplication via appropriate arrangement of tensors into matrices~\cite{chellapilla2006high}.

Remaining network operators such as average and max pooling, depth wise 2D convolution, unbatched linear layers, bias, scale and various nonlinearities reduce to either pointwise operations, vector vector operations or vector matrix operations.  These operations have a lower arithmetic intensity and / or fewer parameters and tend to contribute substantially less to the complexity of common network designs.

Binarizing both the inputs and weights is a method for drastically reducing the implementation complexity of CNN style 2D convolution relative to variants with 32 bit float (common training precision) or 8 bit fixed point (common inference precision) inputs and weights.  While binarization benefits the implementation complexity, it typically results in a model with lower accuracy.  As such, there is an interest in the design and training of high accuracy CNNs which use fully binary CNN style 2D convolution.

This paper:

\begin{itemize}
	\setlength{\itemsep}{1pt}
	\setlength{\parskip}{0pt}
	\setlength{\parsep}{0pt}
	\item Describes a network structure built from a common building block where all CNN style 2D convolution operations are implemented with binary inputs and binary weights; this includes the network stem and down sampling portions of the network
	\item Provides the outline of a constructive universal function approximator proof for binary neural networks which is consistent with the building block design
	\item Uses an existing 2 step training method to achieve 71.24\% top 1 accuracy on the 2012 ImageNet validation set; to the best of the authors' knowledge, this is the highest accuracy on the 2012 ImageNet validation set for a network with all matrix ops fully binary
\end{itemize}

%
%

\section{Related work}

There are a variety of approaches to minimizing CNN complexity while achieving a high level of accuracy.  In practice, implementation complexity is a joint function of network design, software mapping and hardware architecture.  As such, consider the network designs mentioned in this section as a starting point that need to be further considered in the context of a specific software and hardware architecture.


\subsection{Filter sizes}

Initial CNN designs used different filter sizes in the CNN style 2D convolutional layers.  Early attempts at complexity reduction replaced a single layer with a larger filter with multiple layers with 3x3 filters, e.g., replacing a single 5x5 layer with two 3x3 layers.  While not mathematically equivalent, the receptive field size is maintained while the number of parameters and MACs is reduced~\cite{simonyan2014very}.  A somewhat hidden tradeoff, though, is the potential latency and feature map data movement increase, the consequences of which are dependent on the specific hardware architecture.


\subsection{Sparsity}

A logical next step in complexity reduction was the realization that the combined spatial and channel mixing of 3x3 CNN style 2D convolution can be achieved with a depth wise 3x3 2D convolution layer for spatial mixing followed by a separable 1x1 CNN style 2D convolution layer for channel mixing.  As before, while not mathematically equivalent, spatial and channel mixing are still enabled and the parameters and MACs are reduced~\cite{howard2017mobilenets}.

Depth wise 2D convolution is an extreme form of feature map grouping which can be viewed as structured sparsity where only a subset of input channels are connected to each output channel~\cite{xie2017aggregated}.  Other forms of sparsity exist beyond grouped convolution and can be used to reduce complexity.  The more random the sparsity, the more opportunity there is for accuracy vs complexity optimization, but there are also more challenges from an implementation perspective.


\subsection{Building block design}

The above approaches are typically part of the design of building blocks used to construct networks.  The optimal specification of building blocks in terms of input size, channel width and network depth is a key component of efficient network design.  Historically, channels doubled at each down sampling level and building blocks at level 4 were repeated the most as they provided a good mix of receptive field size increase while not increasing the number of parameters as much as level 5.  Recently, large experiments have been used to determine optimal channel width increases and building block repeat patterns for a variety of network complexity design points~\cite{radosavovic2020designing,tan2019efficientnet}.


\subsection{Quantization}

Quantization can be applied to all of the afore mentioned methods to reduce the precision at which computation is performed.  While quantization does not change the theoretical complexity of computations, it has a large effect on the practical implementation complexity.

Memory and data movement, assuming that size changes do not change memory hierarchy locations, scale with the number of bits: if the number of bits per element increases by 2x the memory size increases by 2x.  Integer addition and comparison operations scale with the number of bits used to represent both inputs: if the number of bits in both operands increases by 2x the implementation complexity increases by $\sim$ 2x.

Integer multiplication, however, scales with the number of bits used to represent each input: if the number of bits in both operands increases by 2x the implementation complexity increases by $\sim$ 4x.  As multiplication is more complex from an implementation perspective than addition or comparison, the importance of the precision in layers with lots of multiplications is critical.  Hence the focus on the quantization of the CNN style 2D convolution operation.

Historically, CNN training is done at 32 bit float precision with 24 bits for the sign and mantissa (resolution) and 8 bits for the exponent (range).  Networks can be relatively easily quantized from 32 bit float to bfloat16, with 8 bits for the sign and mantissa and 8 bits for the exponent, as networks tend to be less sensitive to resolution above some threshold and the range stays the same.  The current sweet spot for practical CNN inference is 8 bit fixed point, ideally with range optimization and fixed point quantization built into some portion of the training.  Other combinations of precision for the inputs and weights have been used beyond those mentioned above.


\subsection{Binary CNN design}

The most efficient potential quantization is the extreme end point:  1 bit for the inputs and 1 bit for the weights.  An abbreviated arch of papers using binary CNN style 2D convolution includes:

\begin{itemize}
	\setlength{\itemsep}{1pt}
	\setlength{\parskip}{0pt}
	\setlength{\parsep}{0pt}
	\item XNOR-Net used binary CNN style 2D convolution with output scaling as an approximation of real CNN style 2D convolution~\cite{rastegari2016xnor}; extensions such as XNOR-Net++ looked at different options for output scaling to improve accuracy~\cite{bulat2019xnor}
	\item ABC-Net improved the binary CNN style 2D convolution approximation of real CNN style 2D convolution via a 2 level parallel structure~\cite{lin2017towards}; different branches in the parallel structure used different biases before the sign function to create different binarizations
	\item Bi-Real Net, consistent with the data processing inequality, introduced real identity connections around binary CNN style 2D convolution to minimize information loss~\cite{liu2018bi}; most all high performing binary CNNs have since adopted this strategy as it provides a good tradeoff of accuracy vs complexity for a range of cases
	\item Group-Net considered different serial and parallel combinations of binary CNN style 2D convolution with identity connections~\cite{zhuang2019structured}; philosophically, similar to how ABC-Net generalized XNOR-Net, Group-Net generalized Bi-Real Net
	\item MeliusNet focused on building a better building block via a DenseNet style addition of channels followed by an improvement of those new channels~\cite{bulat2019improved}; however, both the network stem and down sampling stages remained real valued
	\item ReActNet modified the MobileNetV1 structure via making the depth wise separable convolution operations binary 3x3 CNN style 2D convolution (vs depth wise 2D convolution) followed by binary 1x1 CNN style 2D convolution~\cite{liu2020reactnet}; it also introduced trainable biases before the sign and activation function to allow for learnable binarizations and included a binary down sampling stage with channel replication; however, the network stem remained real valued
\end{itemize}


\subsection{Binary CNN training}

The not so hidden challenge of working with binary CNNs is training.  Specifically, almost everywhere there's no gradient propagation through sign functions during automatic differentiation with reverse mode accumulation.  Approaches for addressing this include using real teacher networks and feature map and / or output distribution matching, different functions in the forward and backward paths and / or gradually converging from a real network to a binary network.

%
%

\section{Theory}

A 3 layer real neural network can approximate arbitrarily closely any continuous function on a compact subset of $\mathbb{R}^K$~\cite{cybenko1989approximation}.  The universal function approximator property of neural networks underlies their successful application to a wide variety of problems.  As such, it's useful to understand if binary neural networks maintain this same property.


\subsection{Proof outline}

\begin{figure*}[t]
\begin{center}
	\includegraphics[width=\linewidth]{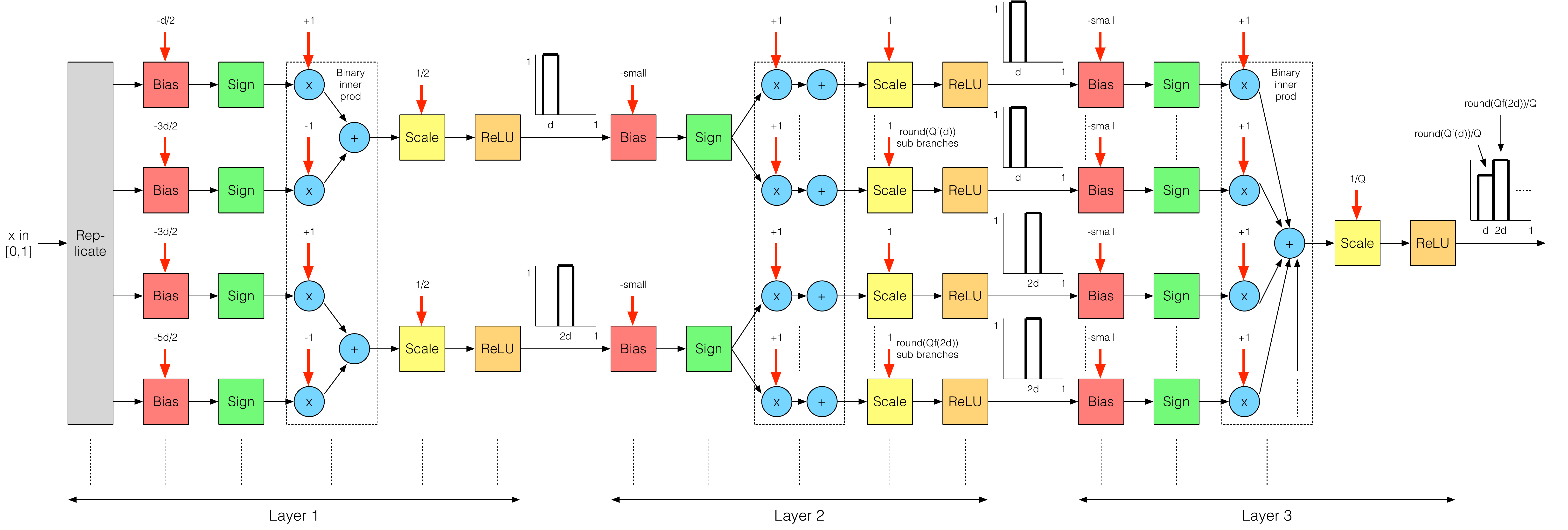}
\end{center}
	\caption{A 3 layer binary neural network that is a universal function approximator.}
	\label{fig:ufa}
\end{figure*}

The following is the outline of a constructive proof, reminiscent of the Reimann integral, showing that a 3 layer binary neural network with a particular building block structure maintains the universal function approximator property.  The proof outline will initially consider the simpler case of showing that a binary neural network can approximate arbitrarily closely a compact function that maps a scalar input $x \in [0, 1]$ to a continuous scalar output $0 \leq f(x) < \infty$, then remove restrictions.

Consider the 3 layer binary neural network structure shown in Figure~\ref{fig:ufa} with input x swept from 0 to 1.

Layer 1 uses $2(1/d - 1)$ input branches to create $1/d - 1$ branches, each generating a rectangle of height $1$ and width $d$ centered at $d, 2d, \ldots, 1 - d$, respectively.  For convenience this omits a $1/2$ width rectangle at the beginning and end of the unit interval which can be added if more formality is desired.

Layer 2 expands the branch centered at d to $\text{round}(Q f(d))$ sub branches, the branch centered at $2d$ to $\text{round}(Q f(2d))$ sub branches, $\ldots$, and the branch centered at $1 - d$ to $\text{round}(Q f(1 - d))$ sub branches.  Each sub branch generates a rectangle of height $1$ and width $d$ centered at the same location as the branch.

Layer 3 combines all of the sub branches of all of the branches and scales the result to generate the final output.

Let $d \to 0$ to approximate all $x \in [0, 1]$ and $Q \to \infty$ to approximate the corresponding value $f(x)$ to complete the initial portion of the proof outline.

Restrictions in the initial proof outline can be removed as follows:

\begin{itemize}
	\setlength{\itemsep}{1pt}
	\setlength{\parskip}{0pt}
	\setlength{\parsep}{0pt}
	\item For $x \in [x_\text{min}, x_\text{max}]$ where $x_\text{min}$ and $x_\text{max}$ are finite, choose branch locations $x_\text{min} + d, x_\text{min} + 2d, \ldots, x_\text{max} - d$ to uniformly tile the full interval
	\item For $f(x)$ with a countable number of discontinuities, center an additional branch at each discontinuity to handle the jump there exactly
	\item For $f_\text{min} \leq f(x) \leq f_\text{max}$ where $x_\text{min}$ and $x_\text{max}$ are finite and each can be negative, $0$, or positive, approximate the non negative function $f(x) + f_\text{min}$ then include a bias term at the end from an additional layer as $-f_\text{min}$
	\item For a vector input $x \in \mathbb{R}^K$ replicate layer 1 for each element of the vector, in layer 2 generate $\text{round}(Q(f(d)))$ $K$ dimensional rectangles for each combination of layer 1 branches and add the $K$ dimensional rectangles together in layer 3
	\item For a vector output $f(x) \in \mathbb{R}^M$ replicate layer 3 for each of the $M$ outputs
\end{itemize}

This completes the outline of the constructive proof that the 3 layer binary neural network structure in Figure~\ref{fig:ufa} is a universal function approximator.


\subsection{Implications}

In the corresponding real valued constructive proof, $d \to 0$ implies more branches.  From a practical perspective, more branches are needed in regions of higher function variation and fewer branches are needed in regions of less function variation.  This matches what would be expected by traditional sampling theory.

The binary valued constructive proof adds the additional constraint $Q \to \infty$, which implies even more branches are needed in the binary neural network case (being somewhat loose with the concept of infinity).  From a practical perspective, to maintain a given accuracy in the function approximation more sub branches are needed for functions with large range, fewer sub branches are needed by functions with small range.  This matches what would be expected by traditional quantization theory, and also implies that a binary neural network will tend to have more channels than a similar accuracy real valued neural network.

%
%

\section{Design}

This section describes the design of an ImageNet image classification CNN where all operations that can be lowered to matrix matrix multiplication are fully binary.


\subsection{Network structure}

\begin{figure}[b]
\begin{center}
	\includegraphics[width=0.5\linewidth]{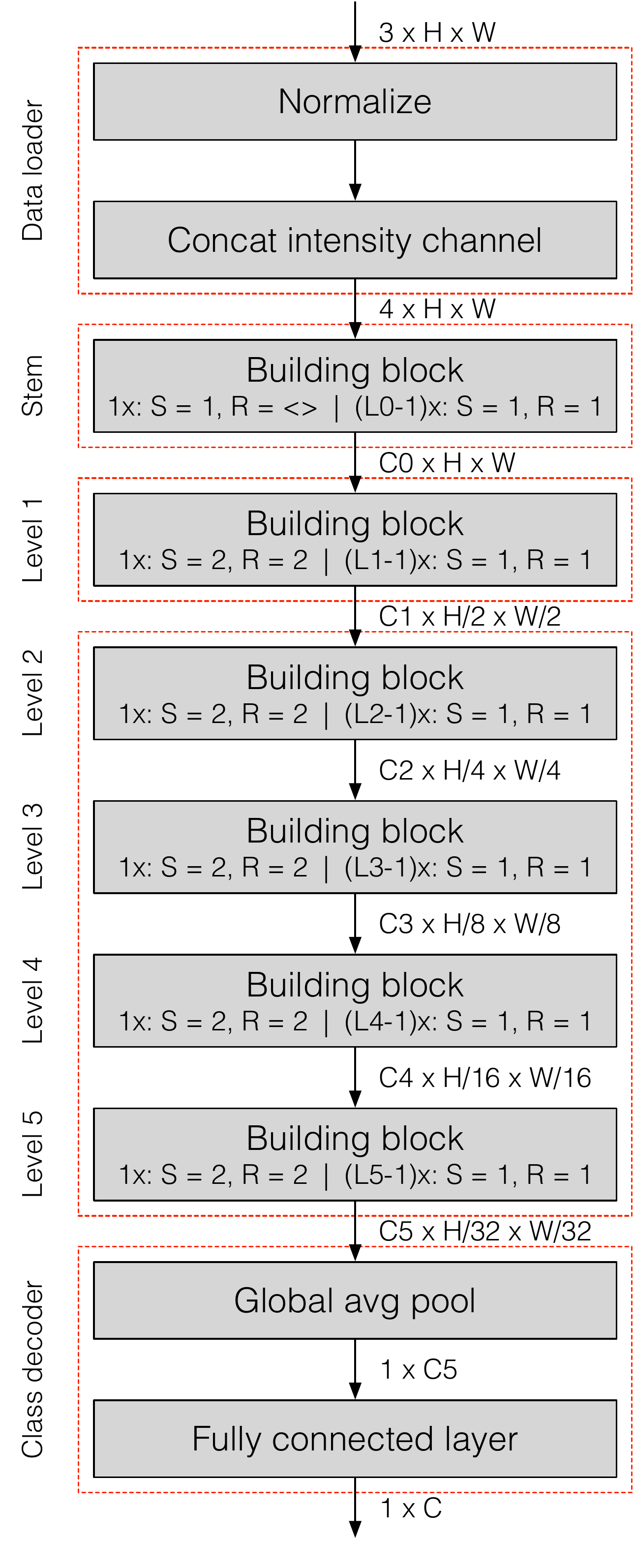}
\end{center}
   \caption{An ImageNet network structure based on a common building block.  Each of the 5 levels has 1 common building block configured for channel replication and down sampling followed by Lx - 1 common building blocks with no channel replication or down sampling, where Lx is the total number of common building blocks in level x.  The stem uses the common building block configured for channel replication only (no down sampling), the selection of the replication factor here determines the number of channels for the remainder of the network.  The decoder uses global average pooling to spatially aggregate features and abstract away the input image size followed by a fully connected layer to classify.}
	 \label{fig:imagenet}
\end{figure}

The ImageNet image classification CNN used in this paper is shown in Figure~\ref{fig:imagenet}.  While it's a relatively typical parameterized structure with 5 levels of down sampling and multiple repeated building blocks, there are a few features to note.

In the data loader a 4th intensity channel is created from the initial 3 RGB channels.  Additionally, places in the network that increase the number of channels by an integer factor were configured to always choose a power of 2 integer factor.  The result is that the number of channels for each level is a power of 2, a convenient value for hardware implementations to work with.  Note that this is not a requirement.

The stem uses the same building block structure as levels 1 - 5.  While this not unheard of, it's a bit less common than a more traditional option like using real 3x3 CNN style 2D convolution.  However, it was done to keep all CNN style 2D convolutions in the network fully binary.


\subsection{Building block}

The building block includes an identity path and a residual path composed of 1x1, 3x3 and 1x1 convolution modules.

\begin{figure}[t]
\begin{center}
	\includegraphics[width=0.9\linewidth]{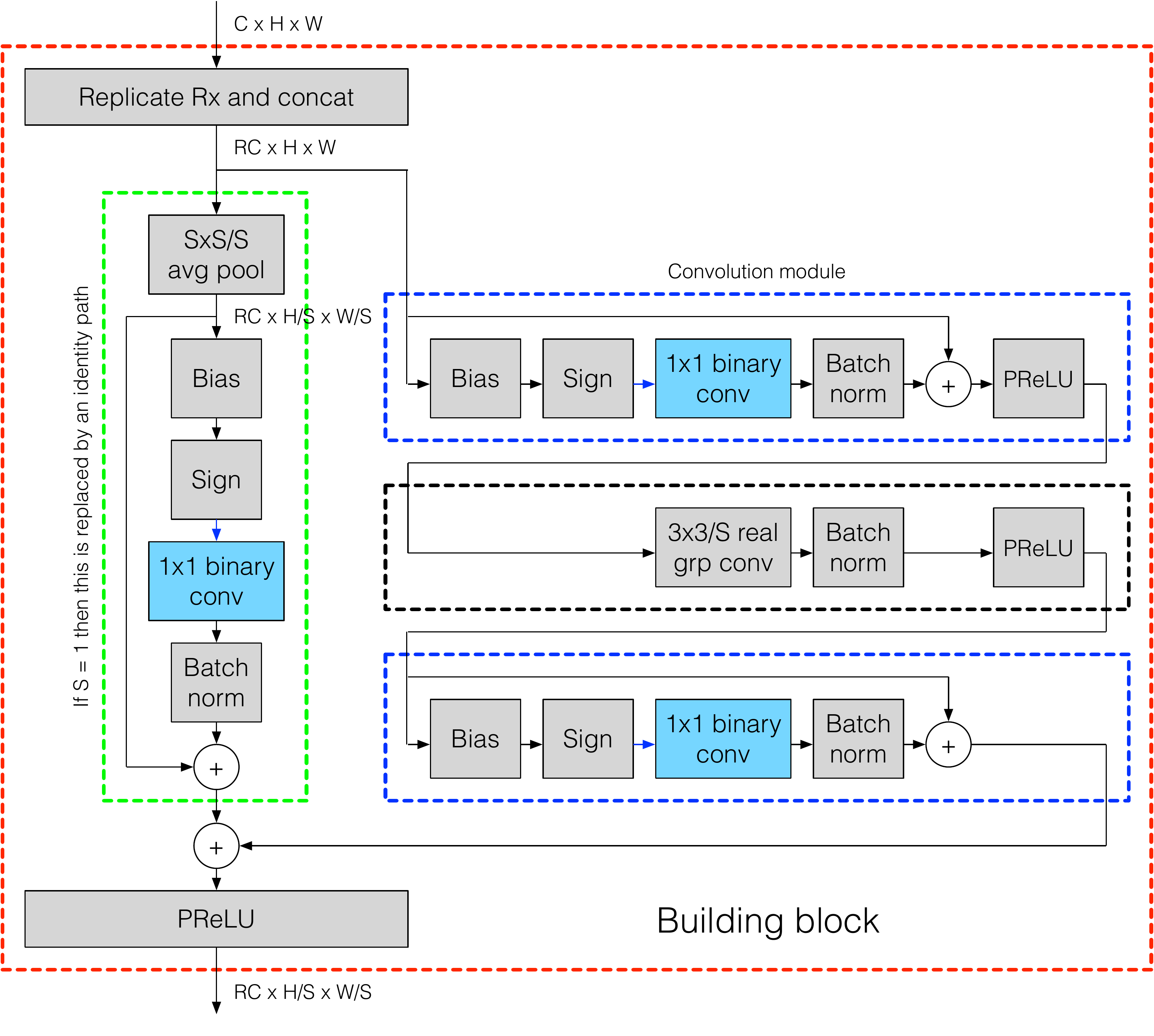}
\end{center}
	 \caption{The stem and level 1 - 5 building block structure capable of integer factor channel increase and integer spatial down sampling; note that all CNN style 2D convolution operations that can be lowered to matrix matrix multiplication are fully binary, the depth wise 3x3 2D convolution that can be lowered to vector matrix multiplication is real.}
	 \label{fig:build_block}
\end{figure}

\subsubsection{Convolution module}

\begin{figure}[t]
\begin{center}
	\includegraphics[width=\linewidth]{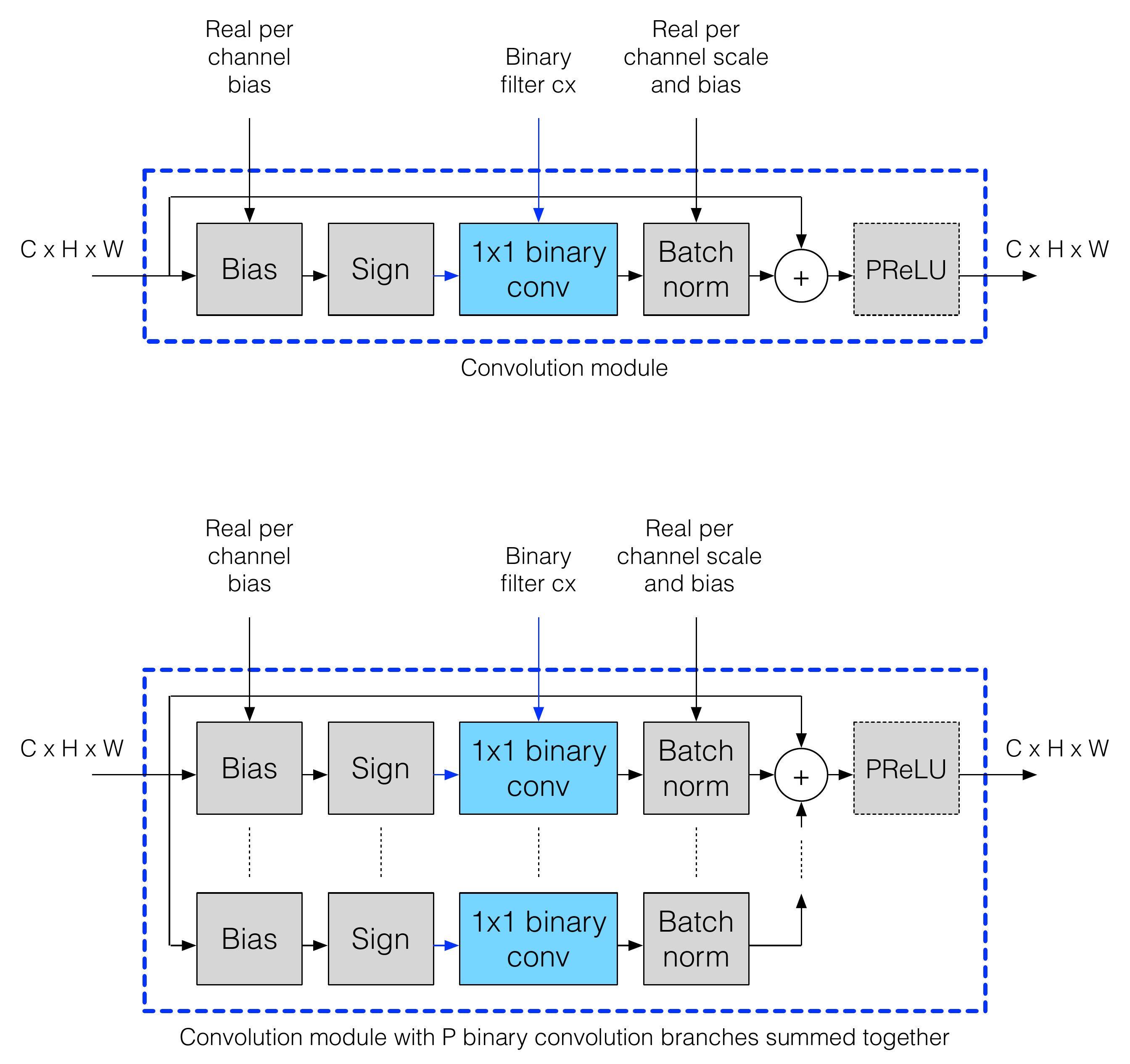}
\end{center}
	 \caption{Convolution module (top) and alternative convolution module with $P$ parallel binary convolution residuals (bottom).}
	 \label{fig:binary_conv_block}
\end{figure}

The 1x1 convolution modules include their own identity path and residual path with learnable bias and sign for the binarization of activations, fully binary 1x1 CNN style 2D convolution for mixing across channels, batch norm for per channel scale and bias, convolution module identity path addition and PReLU.  This follows a Bi-Real Net style approach to the identity connection and a ReActNet style approach to the learnable bias before sign.

Binary 1x1 CNN style 2D convolution implies that each input feature is equally present (+1) or not present (-1) and equally positively (+1) or negatively weighted (-1).  Larger numbers of input and output channels allow more nuance in the values produced by this operation.

The 3x3 convolution module includes real 3x3 depth wise 2D convolution (a vector op) for spatial mixing, batch norm and PReLU.  It's reasonable to ask: Why not use binary 3x3 CNN style 2D convolution for this module?  Using a binary filter for spatial mixing to create a new center value in a 3x3 region as a +-1 combination of the center and neighboring values is not consistent with the typical shape of real filters.  However, successful binary CNNs have made use of binary 3x3 CNN style 2D convolution.  But for the above reason, it was not used here.  On the upside, this operation reduces to a vector op and the number of parameters and MACs is low, so it is in keeping with the theme of implementation complexity reduction.

Inside the binary convolution modules it's possible to use $P$ parallel residual branches, each with a different learned bias, binary weights and batch norm parameters as shown in Figure~\ref{fig:binary_conv_block}.  This is in the same spirit as ABC-Net and Group-Net.

This parallel approach points to a possible alternative for binary $F$x$F$ CNN style 2D convolution.  Using the parallel building block structure in Figure~\ref{fig:binary_conv_block}, let there be $P=F^2$ parallel branches, with an appropriate feature map shift and 0 pad applied before the bias operation of each.  This effectively turns binary $F$x$F$ CNN style 2D convolution into $F^2$ binary 1x1 CNN style 2D convolution operations, where each parallel branch benefits from a real scale in the batch norm before combination to enable learning filter weightings closer to what would be found in a real network.


\subsubsection{Channel increase}

Because of the identity connections around binary CNN style 2D convolution and the fully grouped nature of the real 3x3 depth wise 2D convolution, it's not convenient to increase the number of channels with the convolution operations in the building block.  Instead, when it's necessary to modify the number of channels, channels are simply replicated by an integer factor $R > 1$ at the start of the building block.  In practice, no extra memory is needed for this as the replication can be done by bookkeeping.

Theoretically, $R$ could be chosen as a non integer value larger than $1$ such that not all the channels are replicated.  Likewise, $R$ could be chose as a non integer value less than $1$ with the behavior defined as averaging in the channel dimension.  Channel replication was also used as part of the ReActNet channel increase strategy and is in line with the universal function approximation proof outline.


\subsubsection{Spatial down sampling}

When it's necessary to spatially down sample by a factor of $S$, in a spirit similar to common modifications to ResNet~\cite{he2019bag}, the identity path is replaced by $S$x$S$/$S$ average pooling for spatial mixing followed by a binary 1x1 convolution module for channel mixing.  Likewise, in the residual path, 3x3/$S$ depth wise convolution is used in the real 3x3 convolution module for spatial mixing (followed by channel mixing in the subsequent binary convolution module).  A positive aspect of this strategy is that it allows the identity connections inside the binary 1x1 convolution modules to remain identity connections.


\subsubsection{Connections to real networks}

Connecting with recent real network building block designs, if the binary 1x1 convolution modules were replaced by real 1x1 convolution modules and larger group sizes were used in the real 3x3 convolution module, then this structure would effectively be a RegNetX building block~\cite{radosavovic2020designing}.  Many other network building blocks also use an identity path with variants of a 1-3-1 residual structure initially made famous by ResNet~\cite{he2016deep}.

%
%

\section{Training}

Both 1 step (direct training of binary weights) and 2 step (step 1 training of real weights followed by step 2 binarization and training of binary weights) methods of training were tried.  2 step training provided a slightly higher accuracy, though it's not clear if this is a fundamental to real pre training or just a consequence of the specific hyper parameters used in the training.  Both 1 and 2 step methods borrowed from strategies used by~\cite{liu2020reactnet,liu2018bi,martinez2020training}, the 2 step method is described below.

Note that none of the recent training methods commonly employed to boost ImageNet image classification accuracy for real networks were used~\cite{cubuk2018autoaugment,he2019bag,touvron2019fixing,xie2020adversarial,xu2019accurate,yalniz2019billion,zhang2017mixup}.  Some combination of these would likely improve the reported accuracy results.


\subsection{Step 1: Binary activations real weights}

In step 1 of training, the input to binary convolution was binary but the weights were allowed to take on real values (they will be binarized in step 2).  Step 1 hyper parameters included:

\begin{itemize}
	\setlength{\itemsep}{1pt}
	\setlength{\parskip}{0pt}
	\setlength{\parsep}{0pt}
	\item Batch size 128 (4 GPUs, 32 images per GPU)
	\item Random resized crops to 3 x 224 x 224, random horizontal flip and 0 mean unit variance normalization
	\item Distributional loss with output pmf targets provided by ResNet34
	\item Adam optimizer
	\item 5 epochs of linear warm up followed by 20 epochs of half wave cosine decay
	\item A max learning rate of 0.0005, initial learning rate scale of 0.01x and final learning rate scale of 0.001x
	\item L2 weight decay applied to the convolution parameters with a weighting of 1e-5
\end{itemize}

Step 1 was used to get weights in the ballpark, not to maximize accuracy of binary activations and real weights (only $5 + 20 = 25$ epochs were used).  An alternative weight decay that may degrade the accuracy of step 1 but provide a better starting point for step 2 is described in~\cite{darabi2018regularized}.


\subsection{Step 2: Binary activations binary weights}

Weights were transferred from the last checkpoint in step 1 to initialize step 2.  The real weights for the binary convolution module were used for gradient accumulation in the backward pass, but passed through a sign function to enable fully binary convolution in the forward pass.

Step 2 used the same hyper parameters as step 1 with the following exceptions:

\begin{itemize}
	\setlength{\itemsep}{1pt}
	\setlength{\parskip}{0pt}
	\setlength{\parsep}{0pt}
	\item 5 epochs of linear warm up followed by 100 epochs of half wave cosine decay
	\item No L2 weight decay
\end{itemize}

Typically, weight decay is needed during real network training to improve generalization and achieve high accuracies on ImageNet.  Binarization has a similar regularizing effect and prevents any of the CNN style 2D convolution weights from becoming large.  Clipping of the associated real accumulated weights was not used.


\subsection{The sign function}

\begin{figure}[t]
\begin{center}
	\includegraphics[width=\linewidth]{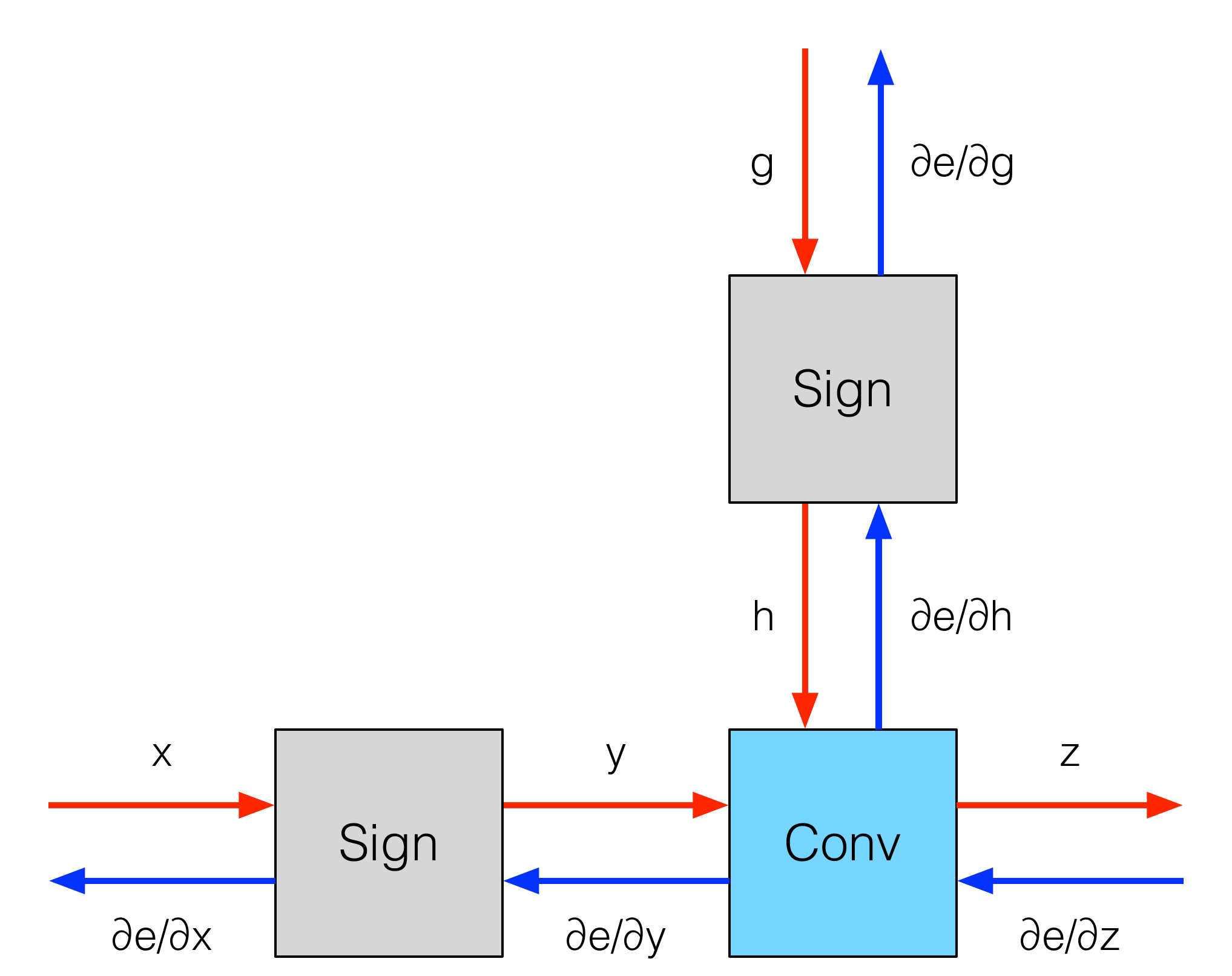}
\end{center}
	 \caption{Zooming in on the (binary) convolution operation to look at the forward flow of weights and activations (red) along with the backward flow of associated error gradients (blue); as both $y$ and $h$ are binary, the convolution operator is a binary convolution operator.}
	 \label{fig:quantization}
\end{figure}

Figure~\ref{fig:quantization} zooms in on the binary convolution operation showing the sign operation in the data path and implicit sign operation applied to real weights to create binary weights.  A key for training binary CNNs is propagating the sensitivity of the error with respect to feature maps and weights through the sign operation as the derivative of the output of the sign operation with respect to the input is 0 almost everywhere.

The training strategy used here followed the strategy in~\cite{liu2020reactnet} with the backward function decoupled from the forward function through the sign operator.  The backward function used to propagate the sensitivity of the error with respect to the activations is

\begin{equation}
	\begin{aligned}
		\partial y / \partial x &= - 2 |x| + 2, & x\in [-1, 1]\\
														&=0, & \text{otherwise}
	\end{aligned}
\end{equation}
which corresponds to (unused) piecewise forward function:

\begin{equation}
	\begin{aligned}
		y &= -1, & x < -1\\
			&= x^2 + 2x, & -1 \leq x < 0\\
			&= -x^2 + 2x, &0\leq x < 1\\
			&= 1, &1 \leq x
	\end{aligned}
\end{equation}
The backward function used to propagate the sensitivity of the error with respect to the weights is:

\begin{equation}
	\begin{aligned}
		\partial y / \partial x &= 1, & x\in [-1, 1]\\
														&=0, & \text{otherwise}
	\end{aligned}
\end{equation}
which corresponds to (unused) forward forward function:

\begin{equation}
	\begin{aligned}
		y &= -1, & x < -1\\
			&= x, & -1 \leq x < 1\\
			&= 1, &1 \leq x
	\end{aligned}
\end{equation}

A complementary method of pushing error information deeper into the network with a teacher network (e.g., step 1 as a teacher for step 2 or a different variant of the network) and forming additional errors via intermediate feature map matching was not used~\cite{ding2019regularizing}.

%
%

\section{Results}


\subsection{Accuracy}

For the network defined in Section 4 configured as shown in Table~\ref{tab:network_config} and using the 2 step training defined in Section 5, the top 1 accuracy on the 2012 ImageNet validation set after 5 + 20 epochs in step 1 was 66.63\% and after 5 + 100 epochs in step 2 was 68.46\%.  After an additional cycle of 5 + 100 epochs in step 2 with the max learning rate decreased to 0.00025 the top 1 accuracy improved to 68.96\%.  Training hyper parameter optimization could likely improve this further.

For the same network with $P = 2$ parallel branches in the convolution module, using the same 2 step 2 cycle training procedure, top 1 accuracy on the 2012 ImageNet validation set improved to 71.24\% after the final step.

\begin{table}[h]
\begin{center}
\begin{tabular}{lllllll}
     \hline
                               & Blocks   & S   & R   & RC     & H/S    & W/S   \\
     \hline
     Stem replicate    & 1   & 1   & 8   & 32   & 224   & 224   \\
     Stem normal      & 0   & 1   & 1   & 32    & 224   & 224   \\
     Level 1 down     & 1   & 2   & 2   & 64    & 112   & 112   \\
     Level 1 normal   & 0   & 1   & 1   & 64    & 112   & 112   \\
     Level 2 down     & 1   & 2   & 2   & 128  & 56     & 56   \\
     Level 2 normal   & 0   & 1   & 1   & 128  & 56    & 56   \\
     Level 3 down     & 1   & 2   & 2   & 256  & 28     & 28   \\
     Level 3 normal   & 1   & 1   & 1   & 256  & 28    & 28   \\
     Level 4 down     & 1   & 2   & 2   & 512   & 14   & 14   \\
     Level 4 normal   & 5   & 1   & 1   & 512   & 14   & 14   \\
     Level 5 down     & 1   & 2   & 2   & 1024 & 7     & 7   \\
     Level 5 normal   & 1   & 1   & 1   & 1024 & 7    & 7   \\
     Global avg pool  & 1   &      &      & 1024 & 1    & 1   \\
     Fully connected  & 1   &      &      & 1000 & 1    & 1   \\
     \hline
\end{tabular}
\end{center}
\caption{Network design configuration.  The feature map size at the level output is RC x H/S x W/S.}
\label{tab:network_config}
\end{table}


\subsection{Operations}

Per the discussion in Section 2, it's not fully appropriate to convert binary operations to equivalent real operations by dividing by a single factor.  The approach taken in this sub section is to simply list the specific operations, which while slightly more cumbersome, is more appropriate for evaluating the complexity on a given implementation (though still not fully acceptable).

The number of parameters and number of operations is listed in Table ~\ref{tab:operation_count} for the network specified in Table~\ref{tab:network_config}.  Note that  $\sim90\%$ of the real parameters are in the class decoder fully connected layer.  Reducing the memory requirements in this layer, potentially via binarization, is a direction for future work.

\begin{table}[h]
\begin{center}
\begin{tabular}{lrr}
	\hline
	Item 				&	$P=1$	&	$P=2$	\\
	\hline
	Binary parameters	&	9.04 e6	&	18.09 e6	\\
	Binary MACs		&	2.41 e9	&	4.83 e9	\\
	Real parameters	&	1.15 e6	&	1.19 e6	\\
	Real MACs		&	0.04 e9	&	0.04 e9	\\
	Real adds			&	62.67 e6	&	109.64 e6	\\
	Real mults		&	19.57 e6	&	35.22 e6	\\
	Sign				&	15.65 e6	&	31.31 e6	\\
	PReLU			&	16.41 e6	&	16.41 e6	\\
	\hline
\end{tabular}
\end{center}
\caption{Parameter and operation counts in the proposed network.}
\label{tab:operation_count}
\end{table}

In Table~\ref{tab:simplified_count} the real MACs, real adds, real mults, sign and PReLU are aggregated into real ops / 2 (to match the units of MACs) with powers chosen to better enable binary to real comparisons.  As the implementation complexity of binary CNN style 2D convolution is small, the implementation complexity of the remaining operations and associated data movement plays a larger role in the ultimate system performance.

\begin{table}[h]
\begin{center}
\begin{tabular}{lrr}
	\hline
	Item 				&	$P=1$	&	$P=2$	\\
	\hline
	Binary parameters	&	9.04 e6	&	18.09 e6	\\
	Binary MACs		&	2.41 e9	&	4.83 e9	\\
	Real parameters	&	1.15 e6	&	1.19 e6	\\
	Real OPs/2		&	0.09 e9	&	0.13 e9	\\
	\hline
\end{tabular}
\end{center}
\caption{Parameters and aggregated number of operations.}
\label{tab:simplified_count}
\end{table}


\subsection{Comparisons}

Table~\ref{tab:comparison1} includes comparisons to 2 recent top performing networks which use mostly binary CNN style 2D convolution: MeliusNet and ReActNet.  The comparisons are grouped to an ImageNet 2012 top 1 validation accuracy of 69.2\% $\pm$ 0.2\% and 71.2\% $\pm$ 0.2\% where Mb is million binary parameters, BGM is binary GMACs, MB is million real parameters and RGO is real GOPs/2.  For reference, the top 1 accuracy of MobileNetV1 is 70.6\%, so the accuracy of these binary networks are in the ballpark of a real network optimized for smaller devices.

\begin{table}[ht]
\begin{center}
\begin{tabular}{lrrrrr}
	\hline
	Network 		&	Acc\%	&	Mb		&	BGM			&	MB			&	RGO			\\
	\hline
	MeliusNet42	&	69.2		&			&	9.69			&				&	0.17			\\
	ReActNetA	&	69.4		&	28.25	&	4.82			&	1.08			&	0.03			\\
	BCNN P=1	&	69.0		&	9.04		&	2.41			&	1.15			&	0.09			\\
	\hline
	MeliusNet59	&	71.0		&			&	18.3--		&				&	0.25			\\
	ReActNetC	&	71.4		&	27.55	&	4.69			&	1.77			&	0.16			\\
	BCNN P=2	&	71.2		&	18.09	&	4.83			&	1.19			&	0.13			\\
	\hline
\end{tabular}
\end{center}
\caption{Comparisons to other binary networks.}
\label{tab:comparison1}
\end{table}

Combining real operations with binary operations via the often used but as previously mentioned lacking rule of thumb normalized OPs = binary OPs / 64 + real OPs and combining real parameters (assuming int8 format) with binary parameters allows for the normalized results in Table~\ref{tab:comparison2}.

\begin{table}[ht]
\begin{center}
\begin{tabular}{lrrrrr}
	\hline
	Network 		&	Acc\%	&	Param MB	&	Norm MOPs/2	\\
	\hline
	MeliusNet42	&	69.2		&	10.1--		&	325.---		\\
	ReActNetA	&	69.4		&	4.61			&	108.40		\\
	BCNN P=1	&	69.0		&	2.28			&	131.28		\\
	\hline
	MeliusNet59	&	71.0		&	17.4--		&	532.---		\\
	ReActNetC	&	71.4		&	5.22			&	232.51		\\
	BCNN P=2	&	71.2		&	3.45			&	208.15		\\
	\hline
\end{tabular}
\end{center}
\caption{Normalized comparisons to other binary networks.}
\label{tab:comparison2}
\end{table}

Around 69.2\% accuracy, BCNN P=1 requires only $\sim$ 50\% of the parameter storage but $\sim$ 20\% more normalized operations than ReActNetA.  Around 71.2\% accuracy, BCNN P=2 requires only $\sim$ 66\% of the parameter storage and $\sim$ 10\% less operations than ReActNetC.  These results were obtained with no efficient frontier of accuracy vs complexity optimization for BCNN.

Note that MeliusNet and ReActNet use real non fully grouped convolutions for the stem and / or transitions.  Doing the same for BCNN could likely be used to improve accuracy, but our goal was binarizing all matrix matrix multiplication operations.

For reference, other well known networks with various parts binarized (and associated top 1 accuracy on the 2012 ImageNet validation set) include BNNs (42.2\%), XNOR-Net (51.2\%), Bi-RealNet-152 (64.5\%) and Real-To-Bin Net (65.4\%).  These networks were omitted from the complexity comparison, however, as they are at lower accuracies.

%
%

\section{Conclusion}

This paper presented a CNN designed for image classification where all operations that can be lowered to matrix matrix multiplication are fully binary.  This was enabled via a common building block design motivated by a constructive universal function approximator proof.  A realization of the proposed network achieved 71.24\% top 1 accuracy on the 2012 ImageNet validation data set.

Logical extensions and next steps include:
\begin{itemize}
	\setlength{\itemsep}{1pt}
	\setlength{\parskip}{0pt}
	\setlength{\parsep}{0pt}
	\item Attempting to improve training via feature map matching with a teacher network, L2 weight regularization during step 1 to $\{-1, 1\}$, alternative real to binary transformations and improved hyper parameter tuning
	\item Finding an efficient frontier of accuracy vs complexity for different network configuration parameters
	\item Experimenting with binary $F$x$F$ CNN style 2D convolution using the parallel branch structure and $P=F^2$ binary 1x1 CNN style 2D convolution operations with additional feature maps shift and 0 pad operations before the bias
	\item Designing an inverted residual building block using the same strategy to reduce real feature map memory in the identity path
	\item Using the binary CNN feature encoder in more complex vision problems with binarized decoders
\end{itemize}

%
%

\clearpage
{\small
\bibliographystyle{ieee_fullname}
\bibliography{bcnn}
}

\end{document}